\documentclass[11pt]{article}

\usepackage[preprint]{acl}
\usepackage{tcolorbox}

\usepackage{times}
\usepackage{latexsym}

\usepackage{url}

\usepackage[T1]{fontenc}

\usepackage[utf8]{inputenc}

\tcbuselibrary{breakable}
\usepackage{microtype}

\usepackage{listings}

\usepackage{inconsolata}

\usepackage{graphicx}
\usepackage{booktabs}
\usepackage{subcaption}

\title{Designing Production-Scale OCR for India: Multilingual and Domain-Specific Systems}

\author{
\textbf{Ali Faraz}, 
\textbf{Raja Kolla},
\textbf{Ashish Kulkarni}, 
\textbf{Shubham Agarwal}\\[10pt]
\textit{Krutrim AI, Bangalore, India}\\[2pt]
\textsuperscript{Contact: \{ali.faraz, raja.kolla, ashish.kulkarni, shubham.agarwal1\}@olakrutrim.com
}
}

\begin{document}
\maketitle
\begin{abstract}

Designing Optical Character Recognition (OCR) systems for India requires balancing linguistic diversity, document heterogeneity, and deployment constraints. In this paper, we study two training strategies for building multilingual OCR systems with Vision–Language Models through the \textit{Chitrapathak} series. We first follow a
popular multimodal
approach, pairing a generic vision encoder with a strong multilingual language model and training the system end-to-end for OCR. Alternatively, we explore fine-tuning an existing OCR model, despite not being trained for the target languages. 
Through extensive evaluation on multilingual Indic OCR benchmarks and deployment-oriented metrics, we find that the second strategy consistently achieves better accuracy–latency trade-offs. Chitrapathak-2 achieves 3-6x speedup over its predecessor with being state-of-the-art (SOTA) in Telugu (6.69 char ANLS) and second best in the rest. 
In addition, we present \textit{Parichay}, an independent OCR model series designed specifically for 9 Indian government documents to extract structured key fields,
achieving 89.8\% Exact Match score with a faster inference. Together, these systems achieve SOTA performance and provide practical guidance for building production-scale OCR pipelines in the Indian context.
\end{abstract}

\section{Introduction}

\begin{figure*}[t]
    \centering
    \includegraphics[width=\textwidth]{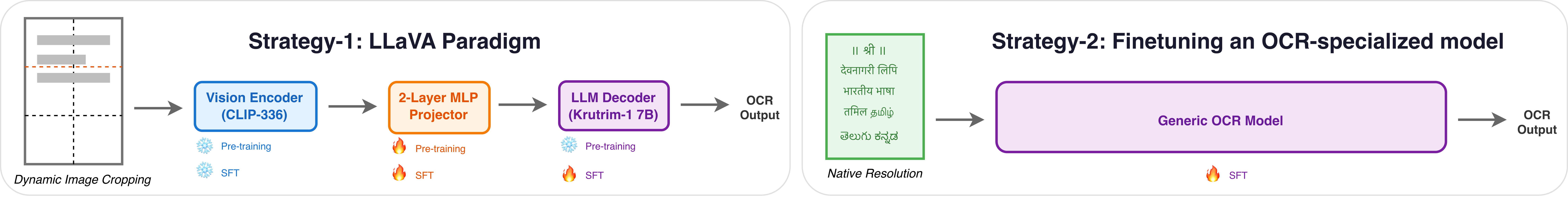}
    \caption{Overview of the two complementary strategies explored in our work in Section \ref{sec:chitrapathak}. We find strategy 2 of finetuning existing model to be data efficient and performing better for multilingual and domain adaptation.}
    \label{fig:Chitrapathak-teaser-image}
\end{figure*}

Optical Character Recognition (OCR) is a foundational component of large-scale document digitization pipelines used across governance, enterprises, and public services in India. Unlike many OCR settings, real-world Indian documents exhibit substantial diversity in scripts, layouts, print quality, and language mixing, often within the same deployment pipeline. At the same time, industrial OCR systems must operate under strict constraints on latency, throughput, cost, and reliability, making system design choices critical in practice.

OCR workloads in India span multiple document regimes, ranging from highly heterogeneous multilingual documents to narrowly scoped but high-volume English government records. These regimes impose different requirements on visual resolution, language modeling, decoding efficiency, and system complexity. As a result, designing OCR systems for India is less about identifying a single best model and more about selecting appropriate design strategies based on document scope and deployment constraints.

Recent Vision-Language Models (VLMs) provide a flexible framework for OCR by directly mapping document images to text, but they admit multiple training strategies. One approach follows a LLaVA-style paradigm \citep{liu2023llava,liu2024improved}, pairing a generic vision encoder with a strong multilingual language model and training the system end-to-end for OCR in multiple stages. An alternative approach involves finetuning an existing VLM-based OCR model for the domain (and languages) of interest. 
Both strategies are reasonable design choices, yet their practical trade-offs for large-scale Indic OCR remain underexplored.

In this work, we study these two strategies (see Figure \ref{fig:Chitrapathak-teaser-image}) through the Indic multilingual OCR series: \textit{Chitrapathak 1 \& 2}.
Through extensive evaluation on multilingual Indic OCR benchmarks and deployment-oriented metrics, we find that the second strategy consistently achieves better accuracy-latency trade-offs, despite originally not trained for Indian languages under study. 

In addition, we also present \textit{Parichay}, an independent OCR model series designed specifically for structured extraction from Indian government documents. Parichay is not evaluated as an alternative to Chitrapathak, but is included as a complementary case study illustrating how strong domain constraints enable simpler architectures and more predictable performance.
Together, Chitrapathak and Parichay\footnote{In Hindi, Chitrapathak is a compound word for `image reader';  Parichay denotes identity} provide practical insights into OCR system design for India, highlighting how training strategy, model specialization, and document scope jointly influence accuracy, efficiency, and deployability. The findings of this paper offer actionable recipe and guidance for practitioners building production-scale OCR pipelines in diverse real-world settings. Our contributions thus are: 

\begin{itemize}

\item We formalize and empirically study two principled approaches for multilingual OCR: LLaVA-style end-to-end training with a strong multilingual language model (Chitrapathak-1), and fine-tuning an existing VLM-based OCR model for the languages under study.
Our work builds a compact OCR system supporting ten Indic languages and English, designed for efficient inference and large-scale deployment.


\item Through extensive evaluation on multilingual Indic OCR benchmarks and system-level metrics, we show that fine-tuning an OCR-specialized model achieves consistently better accuracy–latency trade-offs than end-to-end multilingual training.


\item We also introduce Parichay, an independent OCR model series, paired with a pre-processing rotation module for 9 Indian government documents, achieving SOTA score of 89.8\%, surpassing closed source solutions with a faster inference.

\item By jointly analyzing multilingual and domain-specific OCR systems, we distill actionable lessons on training strategy selection, model specialization, and system design for practitioners building OCR pipelines in real-world Indian settings.

    



\end{itemize}

\section{Related Work}

\textbf{Traditional and Neural OCR Systems.} Classical OCR systems followed multi-stage pipelines with heuristic preprocessing, connected-component segmentation, handcrafted feature extraction, and character-level classification \citep{lebourgeois1992fast, ha1995document, amin2001page, smith2007tesseract}. Modern deep-learning-based OCR systems divide the task of OCR into two major components: Text detection and Text transcription. Text detection has sometimes been formulated as an object detection problem \citep{liao2017textboxes, liao2018textboxes++, liao2018rotation, baek2019character} and sometimes as an instance-segmentation problem \citep{yao2016scene, he2017multi, deng2018pixellink}. For text transcription, many works have combined
convolutional feature extractors with recurrent neural networks \citep{shi2016end, shi2018aster, wang2017gated, luo2019moran}.


\textbf{Transformer-Based and Vision–Language OCR:} With the popularity of the transformer architecture \citep{vaswani2017attention}, there has been a shift towards using transformer-based encoder–decoder models that formulate OCR transcription as direct sequence generation from document images and benefit from large-scale pretraining. TrOCR is a representative example achieving strong results on both printed and handwritten text \citep{li2021trocr}.
Models such as Donut, Dessurt and Nougat use a similar transformer-based encoder-decoder architecture for end-to-end document understanding without an intermediate OCR-only step. \citep{kim2022donut, davis2022end,blecher2023nougat}.
More recently, general-purpose vision-language models (VLMs) connect a vision encoder to a large language model and can be prompted for OCR-like transcription \citep{liu2023llava,chen2022pali,hurst2024gpt,gemini2025flash, bai2025qwen25vltechnicalreport}, while others use a similar paradigm but are specialized for OCR and document understanding \citep{wei2024general, wan2024omniparser, poznanski2025olmocr, lv2023kosmos, wei2025deepseek}. Scaling to high-resolution document images further motivates tiling and cropping strategies, as explored in models such as InternLM-XComposer2-4KHD \citep{dong2024internlmxcomposer2}. In production, deployment efficiency is often governed by inference frameworks such as vLLM and PagedAttention \citep{kwon2023pagedattention}.

\textbf{Indic OCR:} OCR for Indic scripts presents additional challenges due to large character inventories, complex ligatures, typographic variability, and limited high-quality labeled data. Recent open-source systems such as Surya and closed-source systems such as Sarvam vision provide multilingual document OCR with support for several Indic scripts \citep{paruchuri2025surya, sarvamvision2026}.
At the same time, large proprietary multimodal models, including the Google Gemini-2.5 family and OpenAI’s GPT-4o family, offer strong multilingual OCR capabilities in practice \citep{gemini2025flash,hurst2024gpt}. However, much previous work on Indic text focuses on scene-text recognition and benchmark datasets, with comparatively less emphasis on high-fidelity printed document OCR under real deployment constraints such as dense layouts, latency, and throughput \citep{mathew2021benchmarking,gunna2022transfer,lunia2024indicstr12}.
This gap motivates our focus on scalable, deployment-oriented OCR systems for multilingual Indic documents.

\section{Multilingual OCR via Vision–Language Models: Chitrapathak}
\label{sec:chitrapathak}

Chitrapathak is a multilingual OCR system designed to operate across diverse Indic document collections.
We study two distinct training strategies enabled by modern vision-language models, leading to different trade-offs in generality, efficiency, and deployment characteristics.

\subsection{LLaVA-Style End-to-End Multilingual OCR (Chitrapathak-1)}

Based on the success of recent vision-language models and  \textit{Visual instruction tuning} ~\citep{liu2023llava,liu2024improved}, we first experiment with this strategy. We follow a similar architecture as other recent India focused VLMs \citep{khan2025chitrarth, khan-etal-2024-chitranuvad}, however with a focus only on OCR capabilities. 
Chitrapathak-1 follows a LLaVA-style end-to-end training paradigm, where OCR is formulated as an image-to-text generation task similar to a general-purpose vision–language model. The model consists of a vision transformer encoder, a projection MLP, and a multilingual language model decoder. We use CLIP-336 \citep{radford2021learning} as the vision encoder and India-specific Krutrim-1 7B \citep{kallappa2025krutrim} LLM
decoder. Visual embeddings are projected into the language token space and decoded autoregressively. A key limitation here arises from CLIP’s fixed input resolution due to learned absolute positional embeddings. Direct resizing of dense document pages degrades small-text recognition. To mitigate this, we adopt an aspect-ratio-aware tiling strategy inspired by InternLM-XComposer2-4KHD \citep{dong2024internlmxcomposer2}, decomposing each page into a global view and multiple local crops. All crops are resized to the supported resolution, encoded independently, concatenated, passed to the MLP and then to the decoder for transcription. Training proceeds in two stages. During multimodal pretraining, only the projection layer is optimized while both encoder and decoder remain frozen, stabilizing learning under noisy OCR supervision by preserving pretrained visual and linguistic representations. During supervised fine-tuning, the projection layer and language model are jointly trained, with the vision encoder kept frozen. 
While the dynamic image cropping 
strategy improves high-resolution OCR quality, 
lack of compatibility with optimized inference stacks like vLLM results in high latency and memory overhead.



\begin{table*}[htbp]
\centering
\small
\setlength{\tabcolsep}{3pt}
\resizebox{\textwidth}{!}{%
\begin{tabular}{lcccccccccccc}
\hline
\textbf{Language} & Hindi & Sanskrit & Bengali & Marathi & Tamil & Telugu & Kannada & Malayalam & Punjabi & Odia \\
\hline
\textbf{Volume} & 19L & 11.5L & 9.16L & 8.5L & 10.1L & 8L & 3.4L & 2.3L & 4.8L & 48K \\
\hline
\end{tabular}%
}
\caption{Language-wise training data volumes used for multilingual OCR Chitrapathak-1 (L = lakh, K = thousand).}
\label{tab:chitrapathak1-data}
\end{table*}


\begin{figure*}[t]
    \centering
    \includegraphics[width=\textwidth]{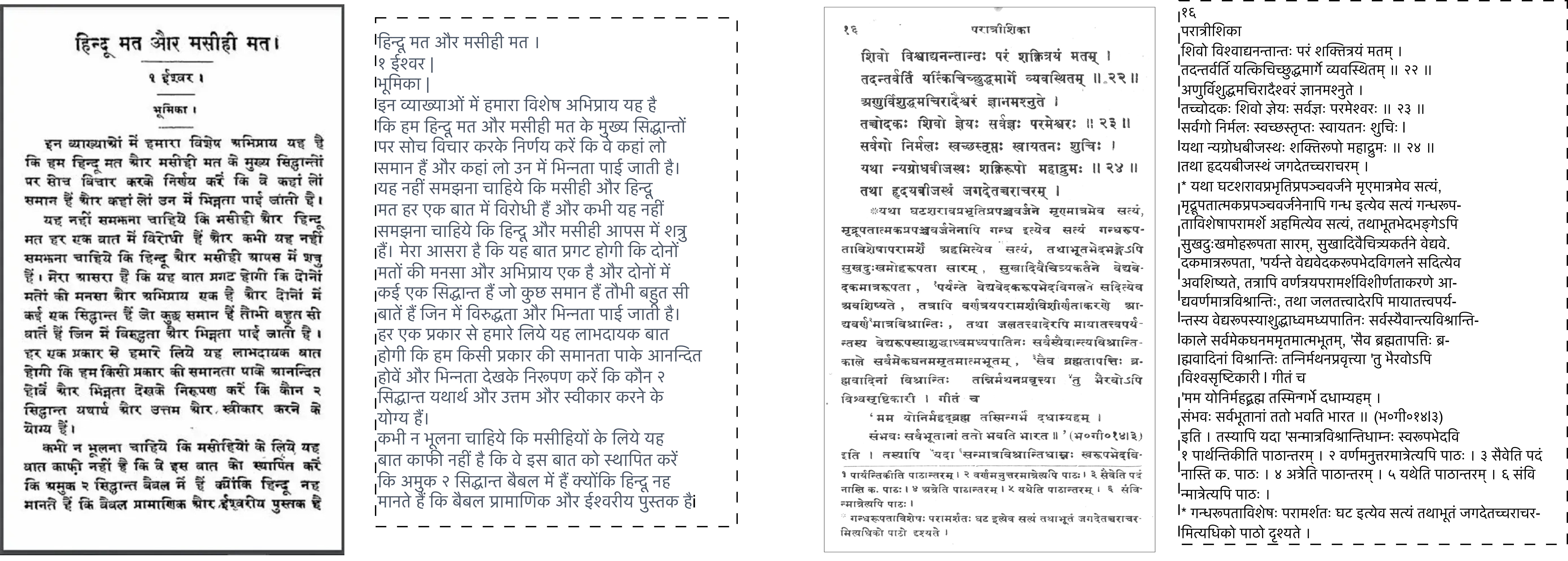}
    \caption{OCR outputs for Hindi (left) and Sanskrit (right) languages from Chitrapathak-2. More examples in Appendix.}
    \label{fig:ocr-examples}
\end{figure*}



\subsection{Fine-Tuning an OCR-Specialized Model for Multilingual OCR (Chitrapathak-2)}

Chitrapathak-2 represents a deployment-oriented redesign guided by efficiency constraints. Here, we fine-tune Nanonets-OCR2-3B \citep{souvik2025nanonets}, built on the Qwen2.5-VL architecture \citep{bai2025qwen25vltechnicalreport}. This backbone uses a native-resolution-capable vision encoder with 2D-RoPE \citep{heo2024ropevit} and windowed attention, eliminating the need for dynamic tiling and allowing direct processing of document images in native resolution within the visual token budget. The model retains the standard vision–language interface: visual tokens are projected via an MLP into a 3B-parameter Qwen-2.5 decoder \citep{bai2025qwen25vltechnicalreport} and decoded autoregressively. Unlike Chitrapathak-1, no additional multimodal pretraining stage is required as the base model is already optimized for OCR-style image-to-text generation. Although the underlying LLM decoder supports the target languages, the base OCR model was not exposed to Indic data during multimodal training. We therefore directly perform supervised fine-tuning on multilingual Indic OCR data to adapt the model to the target scripts and document distributions.
This 
architecture is fully compatible with vLLM, enabling efficient batching, memory management, and token-level scheduling, yielding substantially lower inference latency. 
Figure~\ref{fig:ocr-examples} shows example outputs from Chitrapathak-2.

\begin{table*}[t]
\centering
\scriptsize
\setlength{\tabcolsep}{3pt}
\resizebox{\textwidth}{!}{%
\begin{tabular}{lrrrrrrrrrrrrrrrrrr}
\toprule
\textbf{Model} & \multicolumn{2}{c}{\textbf{Bengali}} & \multicolumn{2}{c}{\textbf{Hindi}} & \multicolumn{2}{c}{\textbf{Kannada}} & \multicolumn{2}{c}{\textbf{Malayalam}} & \multicolumn{2}{c}{\textbf{Marathi}} & \multicolumn{2}{c}{\textbf{Odia}} & \multicolumn{2}{c}{\textbf{Punjabi}} & \multicolumn{2}{c}{\textbf{Tamil}} & \multicolumn{2}{c}{\textbf{Telugu}} \\
\cmidrule(lr){2-3}\cmidrule(lr){4-5}\cmidrule(lr){6-7}\cmidrule(lr){8-9}\cmidrule(lr){10-11}\cmidrule(lr){12-13}\cmidrule(lr){14-15}\cmidrule(lr){16-17}\cmidrule(lr){18-19}
& \textbf{Word} $\downarrow$ & \textbf{Char} $\downarrow$ & \textbf{Word} $\downarrow$ & \textbf{Char} $\downarrow$ & \textbf{Word} $\downarrow$ & \textbf{Char} $\downarrow$ & \textbf{Word} $\downarrow$ & \textbf{Char} $\downarrow$ & \textbf{Word} $\downarrow$ & \textbf{Char} $\downarrow$ & \textbf{Word} $\downarrow$ & \textbf{Char} $\downarrow$ & \textbf{Word} $\downarrow$ & \textbf{Char} $\downarrow$ & \textbf{Word} $\downarrow$ & \textbf{Char} $\downarrow$ & \textbf{Word} $\downarrow$ & \textbf{Char} $\downarrow$ \\
\midrule
LLaMA-4 & 31.52 & 13.21 & 25.73 & 11.91 & 36.90 & 11.17 & 75.50 & 45.75 & 20.94 & 8.05 & 97.51 & 86.78 & 29.77 & 12.68 & 31.36 & 10.79 & 57.07 & 18.72 \\
Gemma-3 27B & 42.15 & 24.41 & 46.47 & 29.50 & 84.22 & 54.24 & 92.06 & 72.64 & 50.40 & 31.06 & 92.67 & 70.72 & 70.88 & 42.65 & 39.52 & 16.51 & 86.76 & 54.14 \\
GPT-4o & 55.51 & 32.68 & 54.62 & 35.54 & 94.33 & 69.79 & 94.67 & 78.47 & 63.44 & 37.93 & 94.61 & 73.46 & 68.88 & 40.71 & 74.35 & 43.39 & 95.97 & 70.08 \\
Nanonets-OCR2-3B & 28.56 & 12.42 & 32.26 & 16.78 & 99.38 & 93.07 & 97.24 & 89.81 & 40.97 & 15.92 & 99.82 & 97.11 & 98.70 & 82.84 & 95.25 & 78.83 & 99.42 & 89.39 \\
Surya OCR & 28.76 & 12.61 & 20.11 & 8.38 & 24.4 & 6.37 & 73.46 & 36.37 & 13.41 & 4.33 & 52.7 & 25.58 & 18 & 7.31 & 25.75 & 7.71 & 51.72 & 16.85 \\
Chitrapathak-1 & 17.14 & 7.03 & 25.55 & 13.74 & 26.24 & 8.78 & 71.97 & 48.19 & 15.68 & 6.09 & 50.72 & 31.62 & 17.70 & 7.87 & 19.25 & 5.81 & 38.79 & 11.00 \\
Chitrapathak-2 & \underline{14.51} & \underline{5.47} & \underline{19.87} & \underline{8.36} & \underline{18.80} & \underline{4.81} & \underline{64.47} & \underline{34.70} & \underline{9.82} & \underline{2.27} & \underline{44.74} & \underline{21.83} & \underline{15.24} & \underline{7.06} & \underline{17.66} & \underline{5.68} & \textbf{31.81} & \textbf{6.69} \\
Gemini-2.5 Flash & \textbf{11.30} & \textbf{4.04} & \textbf{16.01} & \textbf{5.88} & \textbf{17.18} & \textbf{4.38} & \textbf{59.64} & \textbf{30.60} & \textbf{8.06} & \textbf{1.79} & \textbf{41.70} & \textbf{18.60} & \textbf{14.56} & \textbf{4.98} & \textbf{15.26} & \textbf{3.01} & \underline{33.32} & \underline{7.16} \\
\bottomrule
\end{tabular}%
}
\caption{IndicVisionBench-OCR performance (ANLS; lower is better) across multiple baselines. ``Word''/``Char'' denote word-/character-level ANLS.}
\label{tab:indicvisionbench-ocr}
\end{table*}

\section{Domain-Specific OCR: Parichay}


In this second case study, we develop a custom model for domain specific OCR where the system is designed to extract \textit{structured key fields} from Indian government documents in English. We consider information extraction from complex identity and vehicle-related documents such as Aadhaar card, PAN card, Registration Certificates, Driving Licences, Insurance certificates, etc. (see Table \ref{tab:data_distribution}) as part of \textit{Parichay} series. Unlike generic OCR pipelines that focus on raw text recognition, Parichay performs document-conditioned field extraction, directly predicting schema-aligned attributes (e.g., name, date of birth, address), enabling downstream automation in real-world document processing workflows. We formulate structured extraction as instruction-conditioned generation: given one or more document images and a schema-specific prompt describing the required fields, the model produces a JSON-formatted output containing the extracted key-value pairs. Based on the findings from multilingual OCR, we adopt Strategy-2 and finetune OCR specialized models for this domain adaption. Here, we experiment with both LoRA style training ~\cite{hu2021lora,houlsby2019parameter} as well as full parameter fine-tuning. Models are trained via supervised instruction-style fine-tuning on a proprietary dataset. In addition, we follow \citet{goswami2025seeingstraight} and integrate a lightweight document-rotation module built on the Phi-3.5 vision encoder \citep{abdin2024phi3} with dynamic cropping, 
to normalize orientation prior to extraction. This preprocessing step significantly improves robustness in real-world deployment.

Our first model Parichay-1 is built on Phi-3.5 Vision Instruct, a 4.2B-parameter multimodal transformer model. To handle dense and heterogeneous layouts, we adopt the same dynamic cropping strategy used in Chitrapathak-1, decomposing each document into a global view and local crops before visual encoding. We also introduce Parichay-2, derived by fine-tuning Nanonets-OCR2-3B on the same dataset and following similar training recipe as Chitrapathak-2.
In contrast to Parichay-1, it is explicitly optimized for compatibility with vLLM and low-latency inference. 

\section{Experiments}

We experiment with Chitrapathak and Parichay models under their respective deployment regimes, as they target distinct OCR workloads. 

\subsection{Dataset and Implementation details}

\textbf{Multilingual Indic OCR datasets.} The training corpus for Chitrapathak-1 consists of more than 7M printed book-page images spanning multiple Indic scripts, collected from public web sources such as online archives. Table~\ref{tab:chitrapathak1-data} summarizes the language-wise data volumes used for Chitrapathak-1. OCR supervision is obtained by running these images through Google Cloud Platform OCR \footnote{\url{https://cloud.google.com/use-cases/ocr}}, acting as noisy ground truth labels. 
The training corpus for Chitrapathak-2 is constructed as a language-wise stratified sample from the full Chitrapathak-1 training corpus, resulting in 1.1M image-text OCR pairs. Refer to Appendix \ref{app:training_details_cp2} for more training details.

\textbf{English government document dataset.} The Parichay models are trained using supervised instruction fine-tuning on a proprietary dataset comprising approximately 21K annotated document samples and evaluated on 5k data samples (see Table \ref{tab:data_distribution}). Each training instance is constructed to reflect real-world usage scenarios: when a document contains multiple pages (e.g., front and back of Aadhaar), all corresponding images are provided jointly as visual inputs, along with a textual prompt specifying the document type and the set of fields to be extracted, exposing the model to significant layout variability. 


\subsection{Metrics}
We benchmark the Indic OCR performance of the Chitrapathak models on IndicVisionBench-OCR \citep{faraz2025indicvisionbench}. 
In addition, we also evaluate the retention of English OCR capabilities on the popular  datasets like Synthdog \citep{kim2022donut} and SROIE \citep{huang2019sroie}. We report word and character level \textit{Average Normalized Levenshtein Distance} (ANLS) \citep{fu2024ocrbench} to be consistent with prior work. For SROIE, we report a free-form version of Percentage Match. See Appendix \ref{app:evaluation_metrics} for details. 


For the Parichay models, we use a proprietary 5K evaluation set.
We evaluate predictions at the field-value level using two complementary metrics: \textit{Exact Match (EM)} and \textit{Percentage Match (PM)}. EM measures strict string equality after standard normalization, while PM provides a softer similarity-based score to account for minor formatting variations, particularly in long fields such as addresses. The \textit{Mean Score} is defined as the average of the two,
computed at the field level and aggregated across documents 
(see Appendix \ref{app:evaluation_metrics}).

\subsection{Results}

We present a comprehensive evaluation of multilingual models across Indic and English OCR benchmarks.
We evaluate against the following models: Gemini-2.5 Flash \citep{gemini2025flash}, GPT-4o \citep{hurst2024gpt}, Gemma-3-27B \citep{team2025gemma}, LLaMA-4-Maverick-17B (LLaMA-4 for brevity) \citep{meta2025llama4}. Nanonets-OCR2-3B \citep{souvik2025nanonets} and Surya OCR \citep{paruchuri2025surya}. We also provide the evaluation results of Parichay models. 

\subsubsection{Chitrapathak}


Chitrapathak-2 consistently outperforms both base Nanonets-OCR2-3B and Chitrapathak-1 across all languages (Table~\ref{tab:indicvisionbench-ocr}). It achieves state-of-the-art performance for Telugu and remains close to Gemini-2.5 Flash on other scripts (average gap: 2.21 word-level and 1.83 character-level ANLS across nine Indic languages). 
However, we also observe degradation on rare Indic scripts that are under-represented in training, and when moving beyond the printed-book domain to other document types such as forms where certain complex layouts remain challenging. In particular, index-page layouts (dense entries, dot leaders, and irregular alignment) and other complicated page structures can lead to ordering errors and missed/merged lines, even when the text is visually legible. See Figure \ref{fig:limitation-examples} for representative 
examples. Chitrapathak-2 also largely retains the English OCR capability of its base model. Table~\ref{tab:english-ocr} summarizes performance on Synthdog and SROIE. 
On the Old Books OCR dataset \footnote{\url{https://github.com/PedroBarcha/old-books-dataset}} (Table~\ref{tab:oldbooks-ocr}), Chitrapathak-2 remains competitive with both its base model and Gemini-2.5. These results demonstrate that fine-tuning an OCR-specialized backbone yields stronger multilingual generalization.

\paragraph{Latency.}
Table~\ref{tab:latency-avg} reports average 
latency across language groups. Chitrapathak-2 achieves a 3--6$\times$ reduction compared to Chitrapathak-1 and is consistently faster than GPT-4o. Decoding latency varies across scripts due to tokenizer granularity. Table~\ref{tab:token-eff-latency} shows token-to-word ratios and projected latency for generating 200 words. English and Hindi exhibit compact tokenization and lower latency, whereas Telugu and Malayalam produce longer token sequences and correspondingly higher decoding time. Across languages, we observe a time-to-first-token (TTFT) of $\sim$125\,ms and an inter-token latency of $\sim$4\,ms/token.

\begin{table}[htbp]
\centering
\resizebox{0.62\columnwidth}{!}{%
\begin{tabular}{lrrr}
\toprule
\textbf{Model} & \textbf{English} & \textbf{Hindi} & \textbf{Others} \\
\midrule
Chitrapathak-1 & 14.38 & 36.42 & 32.87 \\
Chitrapathak-2 & \textbf{3.10} & \textbf{6.59} & \textbf{14.26} \\
GPT-4o & 10.40 & 18.90 & 23.77 \\
\bottomrule
\end{tabular}
}
\caption{Average end-to-end OCR latency (seconds) on our internal evaluation set (18 English, 40 Hindi and 63 images across 9 other Indian languages).}
\label{tab:latency-avg}
\end{table}



\subsubsection{Parichay}

LoRA-based fine-tuning substantially improves Parichay-1 over the base Phi-3.5V Instruct model (Table~\ref{tab:parichay1_lora}). While higher LoRA ranks generally yield improved performance, gains beyond moderate ranks are marginal, with the 512-512 configuration achieving the strongest LoRA results but only modest improvements over smaller configurations. Meanwhile, full fine-tuning provides a significant performance increase, achieving 86.48 Mean Score (EM 82.13\%, PM 90.83\%). Incorporating the rotation module further improves the robustness, increasing the Mean Score to 92.95 (EM 88.7\%, PM 97.2\%). On the other hand, Parichay-2 achieves both higher extraction accuracy and significantly lower latency. When deployed with vLLM, it reaches an average latency of 1.03 seconds per document ($\approx$ 4× speedup over Parichay-1) while achieving the highest Exact Match (89.8\%) when combined with rotation. Table~\ref{tab:parichay_benchmark} compares Parichay variants against their base backbone and a proprietary VLM Gemini-2.5-Flash (Table \ref{tab:appendix_docwise_em} for extended evaluation). Task-specific fine-tuning is critical: the base model performs poorly on structured extraction, while Parichay series
significantly improve performance. Parichay-2 with rotation achieves the highest EM while maintaining substantially lower latency (Table \ref{tab:parichay_latency}).

\begin{table}[h]
\centering
\resizebox{0.7\columnwidth}{!}{%
\begin{tabular}{lc}
\hline
\textbf{Model} & \textbf{Exact Match (EM \%)} \\
\hline
Base Phi-3.5 Vision Instruct & 23.26 \\
Parichay-1 (Full Fine-Tuning) & 82.13 \\
Parichay-1 + Rotation Module & 88.70 \\
Parichay-2 + Rotation Module & \textbf{89.80} \\
Gemini-2.5-flash & 86.00 \\
\hline
\end{tabular}
}
\caption{Exact Match (EM) comparison across Parichay variants and a proprietary VLM. Parichay-2 with rotation achieves the highest extraction accuracy.}
\label{tab:parichay_benchmark}
\end{table}

\section{Discussion}

Across both the case studies, several key
insights emerge: \textit{i).} Initializing from an OCR-specialized model significantly improves data efficiency: Chitrapathak-2, trained on a subset outperformed Chitrapathak-1,  indicating that structural priors for document text reduce adaptation cost as long as the LLM decoder supports the target languages. \textit{ii).} Tokenizer efficiency becomes a dominant latency factor in multilingual OCR, particularly for scripts with high token-to-word ratios such as Malayalam and Telugu
\textit{iii).} In domain-constrained settings, full fine-tuning provides more stable and accurate adaptation than parameter-efficient methods, suggesting that precise visual-text alignment benefits from complete model updates. \textit{iv).} When document schemas are known, as in Parichay, structured extraction pipelines can bypass general-purpose OCR decoding, resulting in up to 4× lower latency and improved predictability. Collectively, these findings emphasize that specialization and infrastructure alignment are central to scalable 
systems.





\section{Conclusion}

In this work, we present two case studies of building production-grade document understanding systems for the Indian linguistic landscape under real-world deployment constraints. We empirically characterize the trade-offs between general VLM adaptation and OCR-specialized fine-tuning under real-world constraints.
While general VLM adaptation demonstrates feasibility for Indic OCR, fine-tuning an OCR-specialized model delivers substantially better accuracy-latency trade-offs.
In domain-constrained settings, Parichay shows that schema-aware fine-tuning further improves extraction accuracy and efficiency. Together, these results provide practical guidance for scalable OCR and document extraction in industrial settings.

\bibliography{custom}

\null
\vfill
\eject

\clearpage
\appendix
\section*{Appendix}

\section{Limitations of the models}
\begin{figure*}[t]
    \centering
    \includegraphics[width=\textwidth]{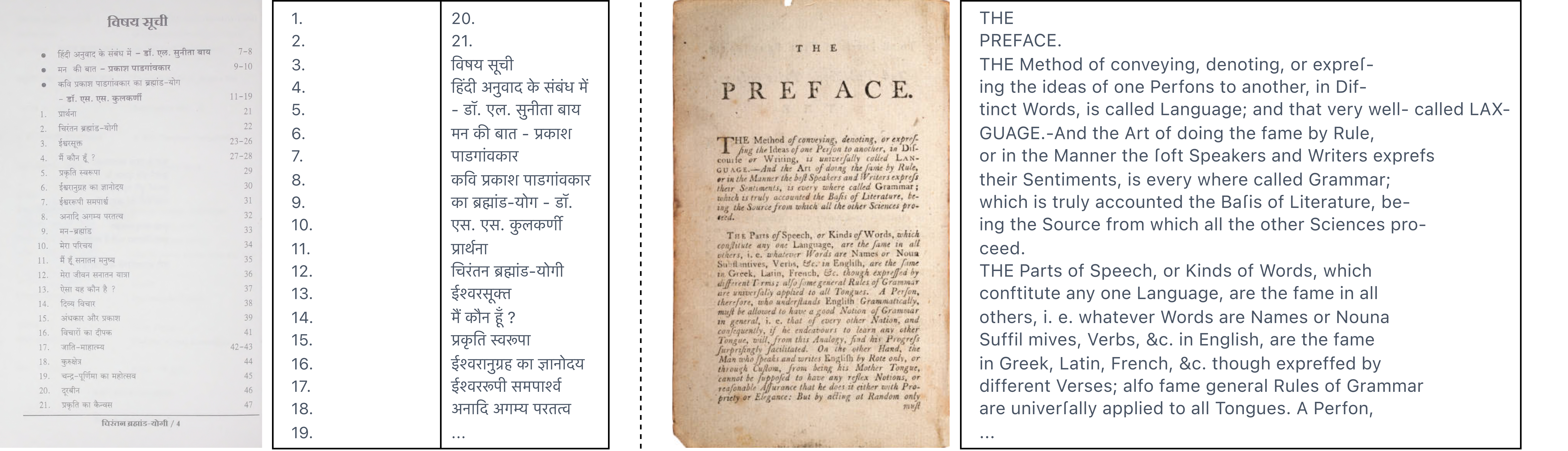}
    \caption{Examples of limitations of Chitrapathak-2. The left image shows an index-page in Hindi while the right image shows a page in English with a rare/old way of writing the letter `s' which the model consistently reads as `f'.}
    \label{fig:limitation-examples}
\end{figure*}

Chitrapathak-2 is designed as an OCR engine for high-fidelity transcription, and is not intended for document intelligence use cases that require structured understanding, field extraction, or reasoning over document content. In particular, given an image, the model returns only the OCR transcription in its default output format. As with most OCR systems, performance can degrade on handwritten text, heavily noisy scans, and low-resolution inputs where character-level evidence is ambiguous. We also observe degradation on rare Indic or English scripts that are under-represented in training, and when moving beyond the printed-book domain to other document types such as forms. Finally, certain complex layouts remain challenging. In particular, index-page layouts (dense entries, dot leaders, and irregular alignment) and other complicated page structures can lead to ordering errors and missed/merged lines, even when the text is visually legible. See figure \ref{fig:limitation-examples} for a few examples.

Parichay is intentionally designed for structured key-field extraction from a predefined set of identity and vehicle-related documents. As a result, its architecture, training data, and prompting strategy are specialized for schema-conditioned outputs, and it does not aim to be a general-purpose document understanding model. Consequently, Parichay exhibits limited performance on tasks outside its target scope, such as free-form OCR, markdown generation, or plain-text extraction. While this specialization enables strong accuracy and low latency for production workflows, extending Parichay to broader document understanding tasks would require additional training data and architectural adaptations.

\section{Training details of Chitrapathak models}
\label{app:training_details_cp2}
\paragraph{Chitrapathak-1.}
Training was performed in two stages: multimodal pretraining and supervised fine-tuning. All experiments used DeepSpeed ZeRO-2 optimization \citep{rajbhandari2020zero} with mixed-precision training (bfloat16) on NVIDIA H100 GPUs, and a maximum sequence length of 4096 tokens.

During multimodal pretraining, only the MLP projection layer was optimized while both the vision encoder (CLIP-ViT-L/14-336) and the language model remained frozen. The model was trained for 1 epoch using an effective batch size of 256 (per-device batch size 2 with gradient accumulation of 16 on 8 GPUs), a learning rate of $1 \times 10^{-3}$, a cosine learning-rate scheduler and a warmup ratio of 0.03.

For supervised fine-tuning, the projection layer and language model decoder were jointly trained for 2 epochs, while keeping the vision encoder frozen. We used an effective batch size of 128 (per-device batch size 1 with gradient accumulation of 16 on 8 GPUs), a learning rate of $2 \times 10^{-5}$, a cosine learning-rate scheduler and a warmup ratio of 0.03.

\paragraph{Chitrapathak-2.}
Training was performed using mixed-precision arithmetic (FP16/bfloat16) and distributed across multiple NVIDIA H100 Nodes with DeepSpeed ZeRO-2 optimization. The model was trained for one epoch with a per-device batch size of 1. We used a learning rate of $1 \times 10^{-5}$, a cosine learning-rate scheduler and a warmup ratio of 0.03.

\section{Evaluation Metrics}
\label{app:evaluation_metrics}

\subsection{Chitrapathak}

For Chitrapathak, we adapt the Percentage Match metric to evaluate structured datasets such as SROIE under a free-form OCR setting. Since Chitrapathak produces unstructured transcription rather than schema-aligned key–value outputs, we evaluate whether each ground-truth field value appears exactly as a substring in the generated OCR text.

Specifically, Percentage Match is defined as the proportion of ground-truth field values that occur verbatim in the free-form OCR output. This formulation differs from the Percentage Match metric used for Parichay, which evaluates structured key–value alignment.

\begin{figure}[h]
\centering
\includegraphics[width=0.4\textwidth]{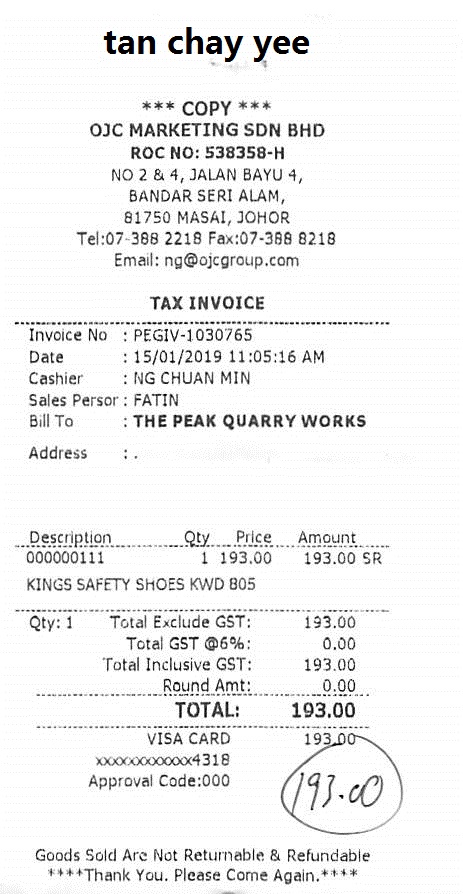}
\caption{Example document image from the SROIE dataset.}
\label{fig:sroie-example}
\end{figure}

Figure~\ref{fig:sroie-example} shows an example image from the SROIE dataset.

For this document, the ground-truth structured annotation is:

\begin{lstlisting}
{
"company": "OJC MARKETING 
SDN BHD",
"date": "15/01/2019",
"address": "NO 2 \& 4, JALAN BAYU
 4, BANDAR SERI ALAM, 81750 
 MASAI, JOHOR",
"total": "193.00"
}
\end{lstlisting}

The corresponding free-form OCR output from Chitrapathak may resemble the following.

\begin{verbatim}
tan chay yee
*** COPY ***
OJC Marketing SDN BHD
ROC NO: 538358-H
...
\end{verbatim}

In this case, the field value \texttt{``OJC MARKETING SDN BHD"} is considered correctly matched because it appears exactly in the OCR output. If even minor deviations occur (e.g., character substitutions or omissions), the field is not counted as a match.

This adaptation is necessary because Chitrapathak is designed for high-fidelity free-form transcription rather than structured extraction. Therefore, for datasets such as SROIE, we convert the structured extraction task into a substring-matching evaluation over free-form OCR outputs to assess English OCR performance.

\subsection{Parichay}
To evaluate structured extraction quality for Parichay, we use a custom field-level evaluation protocol tailored to key-value outputs. For each document instance, the model produces a JSON object containing predicted field names and values. We compare predictions against ground-truth at the \emph{field value} level (conditioned on the same field keys) and report two complementary metrics.

\textbf{Exact Match (EM)} measures strict correctness: a field prediction is counted as correct if the predicted value exactly matches the ground-truth string after standard normalization (e.g., trimming whitespace). Document-level EM is computed by averaging over fields within a document, and dataset-level EM is computed by averaging across all documents.

\textbf{Percentage Match (PM)} provides a softer measure of correctness by quantifying the degree of overlap between the predicted and ground-truth field values. Specifically, PM assigns a similarity score in $[0, 1]$ for each field based on partial matching between the two strings. PM is particularly informative for long or noisy fields (e.g., addresses), where minor formatting variations or OCR artifacts may not reflect semantic extraction failures.

\section{Additional Experiments \& Results}

\subsection{Chitrapathak-1 \& 2}

We report inference efficiency for Chitrapathak-2 across languages (Table~\ref{tab:token-eff-latency}) for a 200-word document input at a resolution of $1024 \times 1024$. 

Total decoding latency is estimated as:

\begin{equation}
\mathrm{Latency}(T) = \mathrm{TTFT} + T \cdot \tau ,
\end{equation}

where $\mathrm{TTFT}$ denotes the time-to-first-token, $T$ is the total number of generated tokens, and $\tau$ represents the average inter-token latency. 

The token count $T$ is derived from the language-specific token-to-word ratio multiplied by 200 words. TTFT and inter-token latency were measured using a standardized latency benchmarking tool GenAI-Perf \footnote{\url{https://tinyurl.com/4e7nh7c8}}.

For the end-to-end latency benchmarking in Table \ref{tab:latency-avg}, Chitrapathak-1 and 2 models were deployed on a single GPU and the inputs were processed sequentially. For GPT-4o as well, we processed the inputs sequentially while using the streaming API.

\begin{figure*}[t]
\centering
\begin{subfigure}[t]{0.49\textwidth}
\centering
\includegraphics[width=\linewidth]{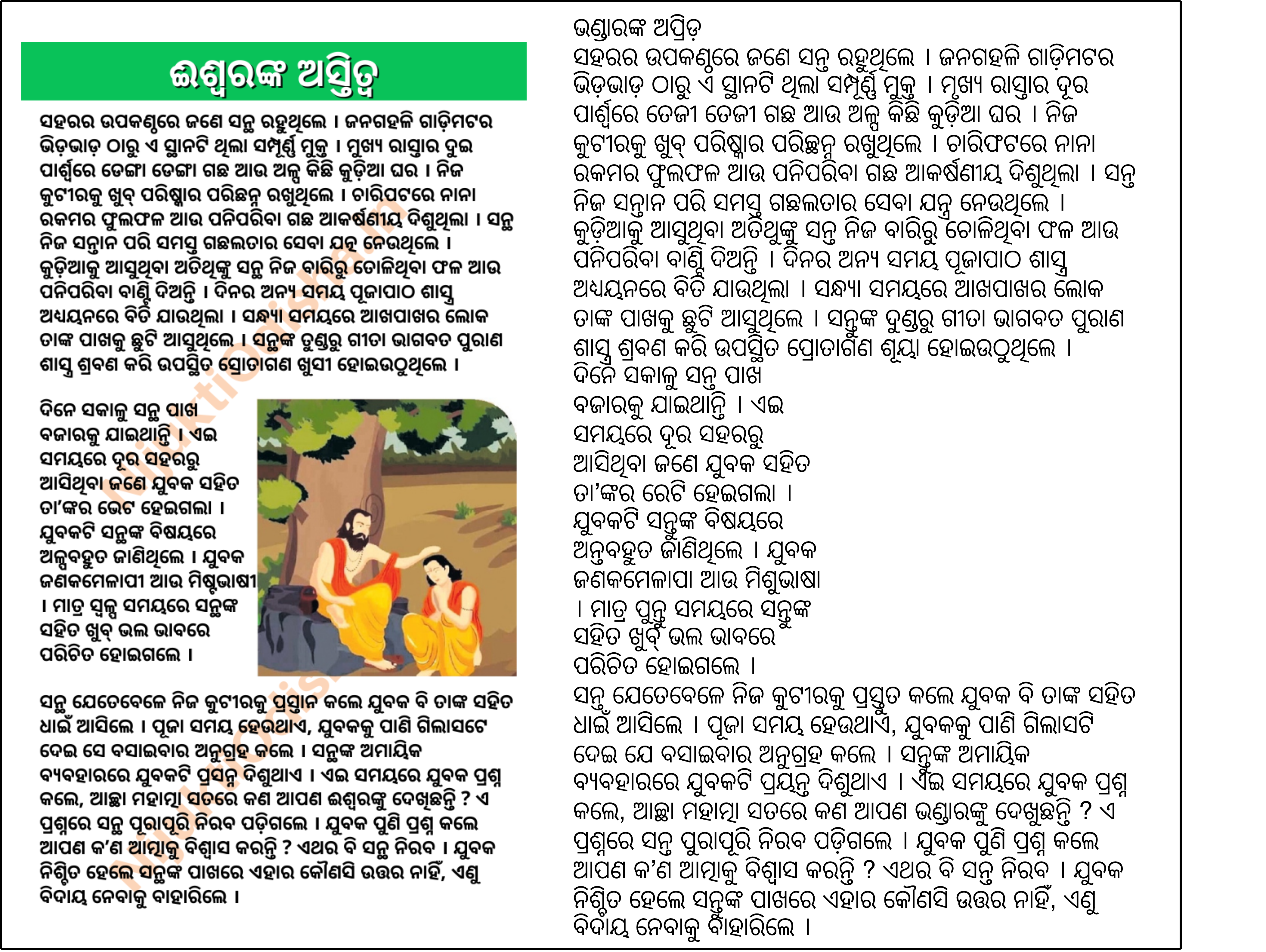}
\caption{Odia}
\end{subfigure}\hfill
\begin{subfigure}[t]{0.49\textwidth}
\centering
\includegraphics[width=\linewidth]{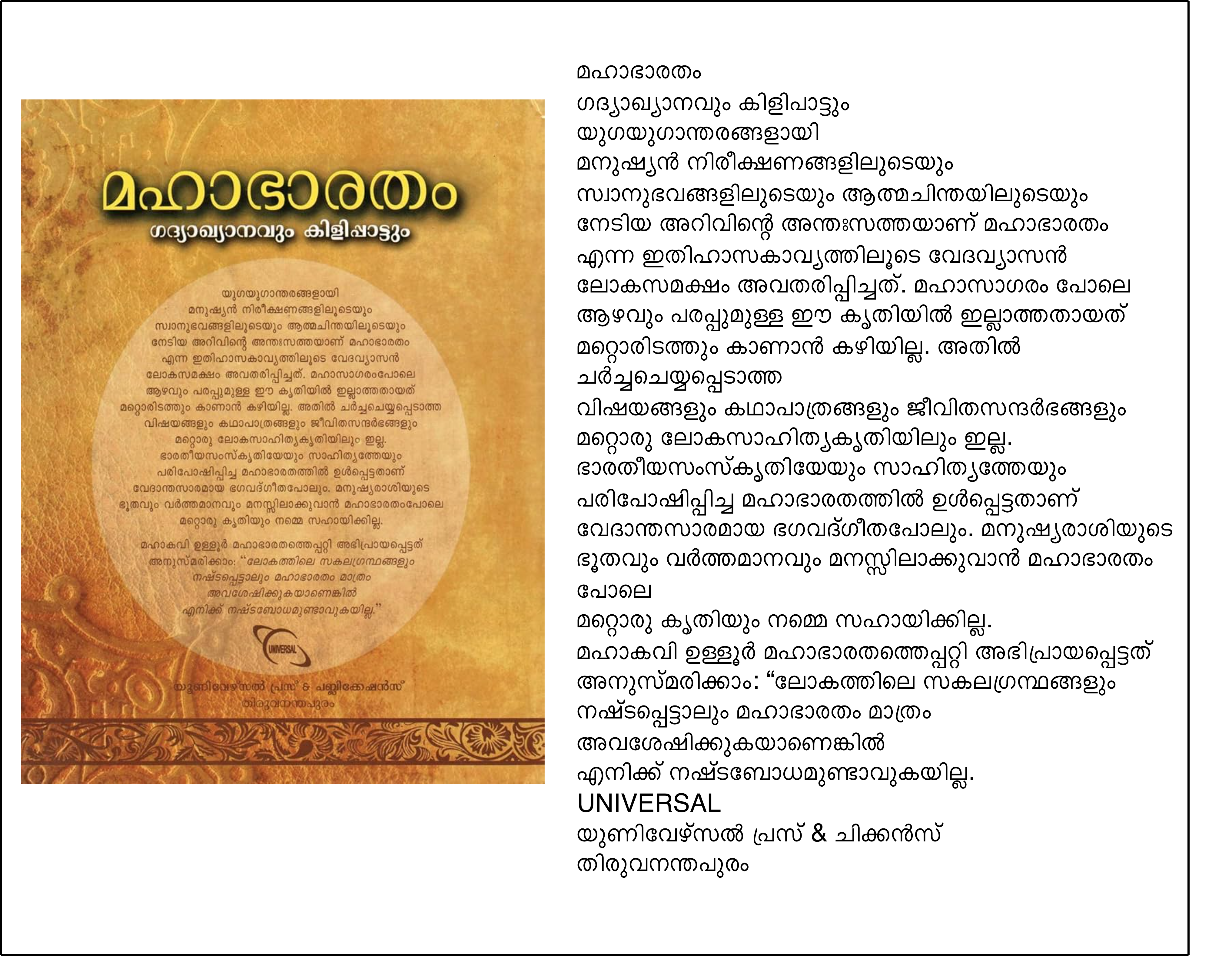}
\caption{Malayalam}
\end{subfigure}
\caption{OCR outputs for Odia (left) and Malayalam (right) languages from Chitrapathak-2}
\label{fig:ocr-examples2}
\end{figure*}


\begin{table}[htbp]
\centering
\small
\setlength{\tabcolsep}{5pt}
\begin{tabular}{lrrr}
\toprule
\textbf{Model} & \multicolumn{2}{c}{\textbf{Synthdog}} & \textbf{SROIE} \\
\cmidrule(lr){2-3}\cmidrule(lr){4-4}
& \textbf{Word} $\downarrow$ & \textbf{Char} $\downarrow$ & \textbf{\%Match} $\uparrow$ \\
\midrule
Gemma-3 27B & 61.56 & 30.29 & 68.37 \\
LLaMA-4 maverick & 29.37 & \underline{14.09} & \underline{70.32} \\
GPT-4o & 82.22 & 73.65 & 36.09 \\
Nanonets-OCR2-3B & \underline{23.90} & \textbf{10.80} & \textbf{72.33} \\
Chitrapathak-2 & 24.90 & 20.20 & 68.95 \\
Gemini-2.5 Flash & \textbf{22.43} & 15.33 & 70.10 \\
\bottomrule
\end{tabular}
\caption{English OCR performance on Synthdog (ANLS; lower is better) and SROIE (\%Match; higher is better).}
\label{tab:english-ocr}
\end{table}

\begin{table}[htbp]
\centering
\small
\setlength{\tabcolsep}{6pt}
\begin{tabular}{lrr}
\toprule
\textbf{Model} & \textbf{Word} $\downarrow$ & \textbf{Char} $\downarrow$ \\
\midrule
Nanonets-OCR2-3B & \textbf{4.33} & 2.96 \\
Chitrapathak-2 & 4.49 & 1.89 \\
Gemini-2.5 & 4.36 & \textbf{1.86} \\
\bottomrule
\end{tabular}
\caption{English OCR performance on the Old Books OCR dataset (ANLS; lower is better).}
\label{tab:oldbooks-ocr}
\end{table}

\begin{table*}[t]
\centering
\scriptsize
\setlength{\tabcolsep}{4pt}
\resizebox{\textwidth}{!}{%
\begin{tabular}{lrrrrrrrrrr}
\toprule
\textbf{Metric} & \textbf{bn} & \textbf{hi} & \textbf{kn} & \textbf{ml} & \textbf{mr} & \textbf{or} & \textbf{pa} & \textbf{ta} & \textbf{te} & \textbf{en} \\
\midrule
\textbf{Tokens/word} & 5.9 & 4.8 & 11.2 & 12.6 & 6.4 & 11.7 & 6.9 & 9.4 & 13.2 & 1.4 \\
\textbf{Tokens (200 words)} & 1174.8 & 951.4 & 2242.2 & 2514.0 & 1292.4 & 2334.2 & 1387.2 & 1873.6 & 2646.6 & 280.0 \\
\textbf{Latency (200 words, s)} & 4.9 & 4.0 & 9.2 & 10.3 & 5.3 & 9.5 & 5.7 & 7.7 & 10.8 & 1.3 \\
\bottomrule
\end{tabular}%
}
\caption{Token efficiency and projected decoding latency for Chitrapathak-2 across languages, assuming a $\sim$1024$\times$1024 input image. All the metrics are rounded to one decimal place.}
\label{tab:token-eff-latency}
\end{table*}

\subsection{Parichay-1}

\begin{table}[h]
\centering
\small
\begin{tabular}{lccc}
\hline
\textbf{Config} & \textbf{Rank ($r$)} & \textbf{Alpha ($\alpha$)} & \textbf{Mean Score (\%)} \\
\hline
Base & -- & -- & 40.45 \\
LoRA & 128 & 128 & 70.13 \\
LoRA & 128 & 256 & 70.54 \\
LoRA & 128 & 512 & 72.34 \\
LoRA & 16  & 256 & 71.50 \\
LoRA & 256 & 256 & 71.87 \\
LoRA & 256 & 512 & 70.15 \\
LoRA & 32  & 256 & 70.86 \\
LoRA & 32  & 512 & 72.51 \\
LoRA & 512 & 256 & 71.79 \\
LoRA & 512 & 512 & 73.03 \\
LoRA & 64  & 256 & 70.74 \\
LoRA & 64  & 512 & 72.08 \\
LoRA & 8   & 256 & 70.35 \\
\hline
Full & -- & -- & \textbf{86.48} \\
\hline
\end{tabular}
\caption{Mean field-level extraction scores for Parichay-1 across LoRA configurations and full fine-tuning. LoRA is applied to the attention projection matrices ($W_q$, $W_k$, $W_v$, $W_o$). Mean Score is computed as the average of Exact Match (EM) and Percentage Match (PM).}
\label{tab:parichay1_lora}
\end{table}

\paragraph{Document-wise Benchmarking.}
Table~\ref{tab:appendix_docwise_em} reports document-wise Exact Match (EM) scores across Parichay variants and multiple baseline systems. Parichay-1 with rotation achieves the highest overall EM of 81.42\%, outperforming all other evaluated models on average across document types. Gemini-2.5-flash follows closely with 79.20\%, while Phi4 with rotation reaches 75.04\%. Among open-source baselines, Llama-4 Maverick-17B attains 71.57\%, whereas Azure+Mistral7B, Nanonets-OCR-s, and Gemma-3-27B-IT achieve 61.09\%, 57.15\%, and 57.94\% respectively. Traditional OCR pipelines combined with language models (DocTR+Mistral7B and Tesseract+Mistral7B) perform substantially worse, with EM scores of 37.09\% and 21.22\%, highlighting the limitations of modular OCR+LLM approaches for structured extraction. Overall, these results demonstrate that task-specific VLM fine-tuning, combined with lightweight preprocessing such as document rotation, delivers significant gains over both generic VLMs and conventional OCR-based pipelines.

\paragraph{Latency Benchmark.}
Table~\ref{tab:parichay_latency} reports the average end-to-end inference latency per document on NVIDIA H100 GPUs.

\begin{table}[htbp]
\centering
\small
\begin{tabular}{lcc}
\hline
\textbf{Model} & \textbf{Inference Engine} & \textbf{Latency (sec)} \\
\hline
Parichay-1 & HF Transformers (generate) & 4.10 \\
Parichay-2 & vLLM & \textbf{1.03} \\
\hline
\end{tabular}
\caption{Average per-document inference latency on H100 GPUs. Parichay-2 achieves nearly 4$\times$ lower latency compared to Parichay-1 while improving extraction accuracy.}
\label{tab:parichay_latency}
\end{table}

\begin{table}[htbp]
\centering
\small
\begin{tabular}{lcc}
\hline
\textbf{Document Type} & \textbf{Train Samples} & \textbf{Test Samples} \\
\hline
Aadhaar & 5166 & 1630 \\
Cancelled Cheque & 4104 & 1074 \\
Car Fitness & 495 & 148 \\
Car Permit & 984 & 45 \\
Driving License & 811 & 241 \\
Insurance & 5010 & 586 \\
PAN & 3389 & 1142 \\
PUC & 808 & 246 \\
RC & 675 & 167 \\
\hline
\textbf{Total} & \textbf{21442} & \textbf{5279} \\
\hline
\end{tabular}
\caption{Distribution of training and test samples across document types used for Parichay model development and evaluation.}
\label{tab:data_distribution}
\end{table}

\begin{table*}[t]
\centering
\scriptsize
\setlength{\tabcolsep}{3pt}
\begin{tabular}{lccccccccccccc}
\hline
\textbf{Doc Type} &
\textbf{Parichay-1 +} & 
\textbf{Parichay-1} & 
\textbf{Parichay-1+} & 
\textbf{Gemini} & 
\textbf{Phi4 +} & 
\textbf{Llama-4} & 
\textbf{Azure +} & 
\textbf{Nanonets} & 
\textbf{Gemma-3} & 
\textbf{DocTR +} & 
\textbf{Tesseract +} & 
\textbf{Mini} & 
\\

 & \textbf{Rot} &
 &
 \textbf{+ Rot (vLLM)} &
 \textbf{2.5} &
 \textbf{Rot} &
 \textbf{17B} &
 \textbf{Mistral7B} &
 \textbf{OCR-s} &
 \textbf{27B-IT} &
 \textbf{Mistral7B} &
 \textbf{Mistral7B} &
 \textbf{CPM} &
 \\
\hline
Aadhaar & 100.00 & 90.00 & 90.00 & 100.00 & 93.33 & 98.00 & 80.00 & 88.00 & 84.00 & 52.00 & 34.00 & 46.00\\
Cancelled Cheque & 60.00 & 60.00 & 15.00 & 80.00 & 65.00 & 70.00 & 75.00 & 40.00 & 25.00 & 35.00 & 10.00 & 0.00\\
Car Fitness & 90.00 & 65.00 & 75.00 & 85.00 & 90.00 & 55.00 & 80.00 & 55.00 & 65.00 & 40.00 & 25.00 & 0.00\\
Car Permit & 77.50 & 77.50 & 40.00 & 62.50 & 55.83 & 50.83 & 58.33 & 45.83 & 43.33 & 50.83 & 35.00 & 15.00\\
Driving Licence & 85.00 & 70.00 & 90.00 & 90.00 & 83.33 & 85.00 & 60.00 & 85.00 & 75.00 & 35.00 & 15.00 & 0.00\\
Insurance & 75.00 & 70.00 & 50.00 & 65.00 & 77.50 & 57.50 & 35.00 & 22.50 & 42.50 & 27.50 & 12.50 & 0.00\\
PAN & 100.00 & 100.00 & 100.00 & 94.44 & 90.00 & 94.44 & 80.00 & 90.00 & 88.89 & 45.00 & 15.00 & 75.00\\
PUC & 72.22 & 72.22 & 38.00 & 66.67 & 60.00 & 72.22 & 30.00 & 25.00 & 27.78 & 25.00 & 30.00 & 0.00\\
RC & 74.00 & 59.50 & 60.00 & 69.50 & 65.71 & 63.50 & 51.50 & 63.00 & 70.00 & 23.50 & 14.50 & 0.00\\
\hline
\textbf{Grand Total} & \textbf{81.42} & 73.52 & 62.00 & 79.20 & 75.04 & 71.57 & 61.09 & 57.15 & 57.94 & 37.09 & 21.22 & 15.11 \\
\hline
\end{tabular}
\caption{Document-wise Exact Match (EM) scores across Parichay-1 variants and baseline systems. Each document category contains 10 evaluation instances (90 samples in total). Parichay-1 with rotation achieves the highest overall EM. The last column reports the number of samples per document type.}
\label{tab:appendix_docwise_em}
\end{table*}

\begin{tcolorbox}[
breakable,
colback=cyan!10!white,
colframe=cyan!70!black,
title=Prompts used to train Parichay-1 and Parichay-2,
fonttitle=\bfseries,
coltitle=black,
boxrule=0.8pt,
arc=3pt,
left=6pt,
right=6pt,
top=6pt,
bottom=6pt
]

\textbf{Fitness Certificate}

Extract the following information from this car fitness document and return it in JSON format:
- "Expiry date"
- "Engine Number"
- "Registration Number"
- "Chassis Number"
Ensure that each key represents the corresponding field and its value represents the extracted data from the document. If any of the fields are not present, omit them from the output without adding any extra text or information outside of the JSON structure.

\vspace{0.5em}
\textbf{Aadhaar}

Extract the following information from this Aadhaar document and return it in JSON format:
"Person Name"
"State"
"Aadhar Number"
"Pin Code"
"Address"
"Gender"
Ensure that each key represents the corresponding field and its value represents the extracted data from the document. If any of the fields are not present, omit them from the output without adding any extra text or information outside of the JSON structure.

\vspace{0.5em}
\textbf{Cancelled Cheque}

Extract the following information from this cancelled cheque and return it in JSON format:
"Person Name"
"IFS Code"
"Account Number"
Ensure that each key represents the corresponding field and its value represents the extracted data from the document. If any of the fields are not present, omit them from the output without adding any extra text or information outside of the JSON structure.

\vspace{0.5em}
\textbf{Driving Licence}

Extract the following information from this driving license and return it in JSON format:
"Person Name"
"Address"
"Date Of Birth"
"Licence No"
"Pin Code"
"State"
"City Name"
"Licence Date Of Issue"
"Non Transport Date Of Expiry"
Ensure that each key represents the corresponding field and its value represents the extracted data from the document. If any of the fields are not present, omit them from the output without adding any extra text or information outside of the JSON structure.

\vspace{0.5em}
\textbf{PAN Card}

Extract the following information from this PAN card and return it in JSON format:
"Person Name"
"Pan Number"
"DOB"
This document will have two names one is person name and next one is father name. Person name will be above the father name. Ensure that each key represents the corresponding field and its value represents the extracted data from the document. If any of the fields are not present, omit them from the output without adding any extra text or information outside of the JSON structure.

\vspace{0.5em}
\textbf{PUC}

Extract the following information from this pollution estimate test certificate and return it in JSON format:
"I.D No."
"PUC Certificate Number"
"Pucc No"
"Certificate SL. No."
"Serial No."
"Vehicle Registration Number"
"Expiry Date"
Ensure that each key represents the corresponding field and its value represents the extracted data from the document. If any of the fields are not present, omit them from the output without adding any extra text or information outside of the JSON structure.

\vspace{0.5em}
\textbf{Registration Certificate}

Extract the following information from this registration certificate and return it in JSON format:
"Person Name"
"Pin Code"
"Engine Number"
"State"
"Fuel Type"
"Chassis Number"
"Color"
"Regn Validity"
"Date of Regn"
"Model"
"Manufacturer"
"Regn. No "
"Address"
Ensure that each key represents the corresponding field and its value represents the extracted data from the document. If any of the fields are not present, omit them from the output without adding any extra text or information outside of the JSON structure.

\vspace{0.5em}
\textbf{Insurance}

Extract the following information from this insurance document and return it in JSON format:
"Chassis Number"
"Insurance Provider"
"Vehicle Registration Number"
"Insurance valid upto"
"Valid From"
"Policy Number"
"Engine Number"
Ensure that each key represents the corresponding field and its value represents the extracted data from the document. If any of the fields are not present, omit them from the output without adding any extra text or information outside of the JSON structure.

\vspace{0.5em}
\textbf{Permit}

Extract the following information from this permit document and return it in JSON format:
"Permit No"
"Chassis Number"
"Validity of permit from"
"Validity of permit to"
"Engine Number"
"Vehicle Registration Number"
"Region covered"
"Permit Issuing State"
Ensure that each key represents the corresponding field and its value represents the extracted data from the document. If any of the fields are not present, omit them from the output without adding any extra text or information outside of the JSON structure.

\end{tcolorbox}

\end{document}